\documentclass[
    10pt,
    twocolumn
    ]{article}

\usepackage{titling}
\usepackage[utf8]{inputenc}

\usepackage[
    colorlinks=true,
    linkcolor=black,
    citecolor=black,
    filecolor=black,
    urlcolor=black,
    plainpages=false,
    pdfpagelabels,
    breaklinks=true
    ]{hyperref}

\usepackage[
    natbib=true,
    backend=bibtex,
    style=nature,
    citestyle=numeric-comp,
    sorting=none,
    giveninits=true,
    uniquename=init
]{biblatex}
\renewbibmacro*{name:andothers}{
    \ifboolexpr{
    test {\ifnumequal{\value{listcount}}{\value{liststop}}}
    and
    test \ifmorenames
  }
    {\ifnumgreater{\value{liststop}}{1}
       {\finalandcomma}
       {}%
     \andothersdelim\bibstring{andothers}}
    {}}
\usepackage{amsthm,amsmath}
\usepackage{amsfonts, amssymb, amscd}
\usepackage{bbm}
\usepackage[font=footnotesize,labelfont=bf]{caption}
\usepackage{changes}
\usepackage{cuted}
\usepackage{enumitem}
\usepackage[capitalize, nameinlink, poorman]{cleveref}
\newcommand{\YYCleverefInput}[1]{} \YYCleverefInput{main.sed}

\crefname{equation}{Eqn.}{Eqns.}
\Crefname{equation}{Eqn.}{Eqns.}
\crefname{methods}{Methods}{Methods}
\crefname{supplementary}{Supplementary Information}{Supplementary Informations}

\usepackage{geometry}
\usepackage{setspace}
\usepackage{graphicx}
\usepackage{listings}                            %
\usepackage{multicol}
\usepackage{multirow}
\usepackage{lipsum}
\usepackage{placeins}  %
\usepackage{sidecap}
\usepackage[binary-units]{siunitx}
\usepackage{threeparttable}
\usepackage{tikz}
\usepackage{todonotes}

\usepackage{xcolor}
\usepackage{xspace}

\newcommand*\subtxt[1]{_{\textnormal{#1}}}
\DeclareRobustCommand\_{\ifmmode\expandafter\subtxt\else\textunderscore\fi}

\date{}

\begin{document}

\title{\textbf{The Yin-Yang dataset}}

\newcommand{\affilDP}{\textsuperscript{1}}
\newcommand{\affilKIP}{\textsuperscript{2}}
\author{
    \normalsize{
        L. Kriener\affilDP,
        J. Göltz\affilKIP$^{,}$\affilDP,
        M. A. Petrovici\affilDP$^{,}$\affilKIP}\\
    \footnotesize{
        \affilDP
        Department of Physiology, University of Bern, 3012 Bern, Switzerland.
        }\\
    \footnotesize{
        \affilKIP
        Kirchhoff-Institute for Physics, Heidelberg University, 69120 Heidelberg, Germany.
        }\\
}

\maketitle
\begin{abstract}
    The Yin-Yang dataset was developed for research on biologically plausible error backpropagation and deep learning in spiking neural networks.
    It serves as an alternative to classic deep learning datasets, especially in early-stage prototyping scenarios for both network models and hardware platforms, for which it provides several advantages.
    First, it is smaller and therefore faster to learn, thereby being better suited for small-scale exploratory studies in both software simulations and hardware prototypes.
    Second, it exhibits a very clear gap between the accuracies achievable using shallow as compared to deep neural networks.
    Third, it is easily transferable between spatial and temporal input domains, making it interesting for different types of classification scenarios.
\end{abstract}

\section{Introduction}

We introduce the Yin-Yang dataset for learning in hierarchical networks \citep{yygithub}.
It is tailored to the requirements of research on biologically plausible error backpropagation algorithms, learning in spiking neural networks and hierarchical networks on neuromorphic hardware.
These fields typically require small but at the same time not trivially solvable datasets to prototype and test network architectures and learning algorithms.
The datasets commonly used for this purpose are the MNIST and the fashion-MNIST datasets \citep{lecun1998gradient, xiao2017fashion}.
However, these require comparatively large networks, considering the size of the visible layer alone.
Despite this ostensible difficulty, they can nevertheless be classified with high accuracy even by shallow networks or networks without learning in the lower layers.
This is problematic because training a deep network with an imperfect learning algorithm can result in performance indistinguishable from that of a shallow network or a network with plasticity only in the last layer.
Conversely, a test on the MNIST dataset can fail to reveal the inability of the training algorithm to propagate error signals through the network, as the achieved high accuracies obscure the underlying problem.

The Yin-Yang dataset can provide an alternative for these testing and prototyping scenarios as it is solvable by smaller networks, contains fewer samples and most importantly exhibits a large gap between the accuracies reached by shallow or partly fixed networks on the one hand and correctly trained deep networks on the other.
Note that here, we use "deep" in opposition to "shallow", i.e., any network that has latent variables through which errors need to propagate.
We consider a shallow network to be the equivalent of a single-layer perceptron, with only an input layer connected directly to a label layer.

\section{Dataset}

\begin{figure}
    \centering
    \includegraphics[width=0.50\textwidth]{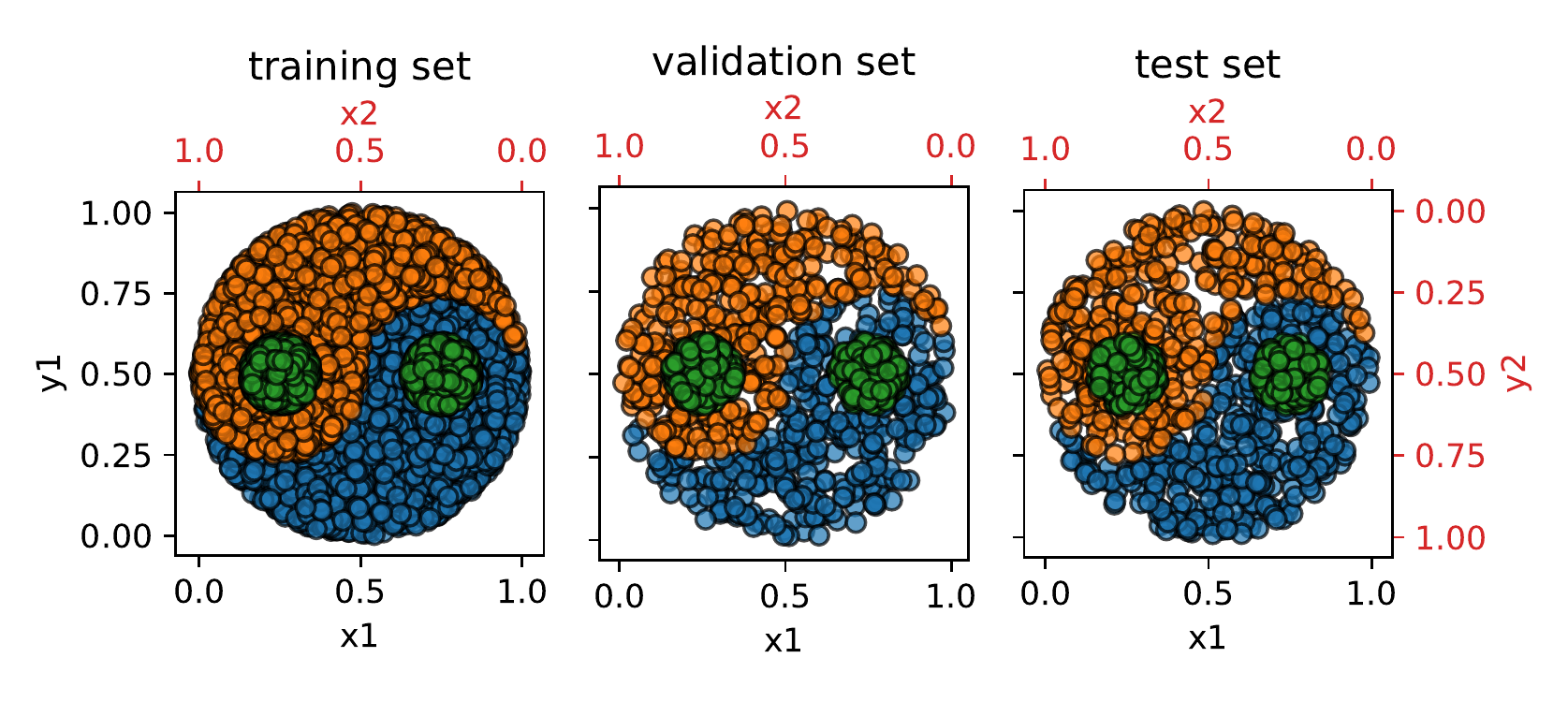}
    \caption{
        \textbf{Training, validation and test dataset.}
        Each dot in the yin-yang symbol represents one sample of the dataset.
        The color of the dot denotes its class (``Yin'', ``Yang'' or ``Dot'').
        This figure was generated using the default settings for random seeds and dataset sizes (5000 samples for the training set and 1000 samples each for the validation and test set).
    }
    \label{fig:datasets}
\end{figure}

\begin{figure}[!ht]
    \centering
    \includegraphics[width=0.48\textwidth]{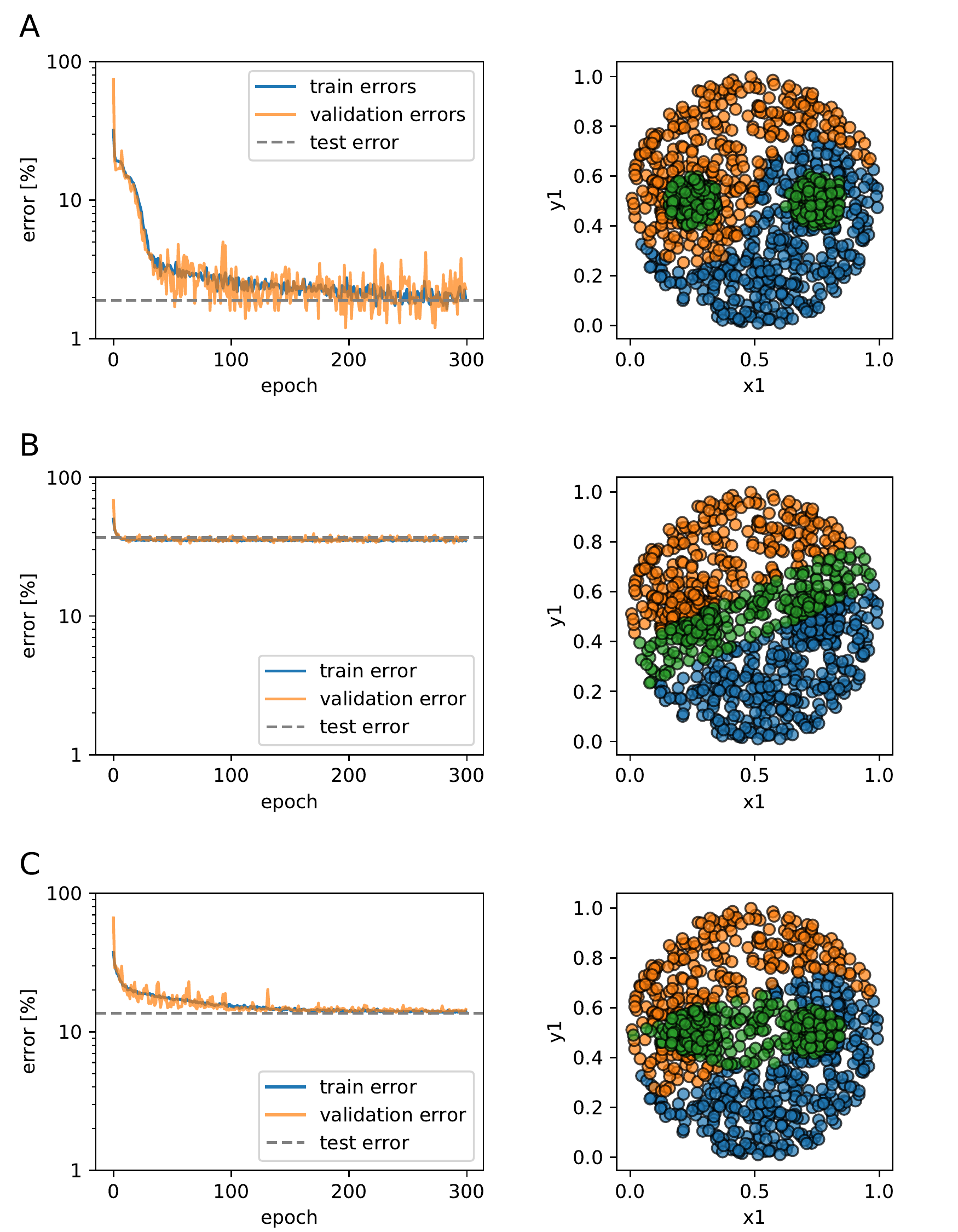}
    \caption{
        \textbf{Comparison of exemplary training results for different network setups.}
        Network parameters are given in \cref{tab:train_params}.
        \textbf{Left column:} Evolution of the validation and training error during training.
        \textbf{Right column:} training result illustrated on the test set.
        \textbf{(A)} Network with one hidden layer and fully functional synaptic plasticity via classical error backpropagation.
        \textbf{(B)} Shallow network.
        \textbf{(C)} Network with one hidden layer and frozen lower weights. result of the deep neural network with frozen weights illustrated on the test set.
    }
    \label{fig:ann_results}
\end{figure}

Each sample in the dataset represents a point in a two-dimensional representation of the yin-yang symbol.
Depending on their location in the symbol the samples are classified into the ``Yin'', ``Yang'' or ``Dot'' class (\cref{fig:datasets}).
Even though the areas in the yin-yang symbol covered by the different classes have different sizes, the dataset is designed to be balanced, which means that all classes are represented by approximately the same amount of samples.
Note that therefore the density of samples is higher in the ``Dot''-class regions, as the combined area of these regions is smaller than that of the others.

The samples are randomly generated using rejection sampling.
The exact version of a generated set of samples is therefore determined by the random seed and dataset size.
This makes it possible to produce multiple dataset versions by providing different random seeds and dataset sizes.
In the default configuration the training set has 5000 samples while the validation and test sets have 1000 samples respectively, each generated with a different random seed.

As can be seen in \cref{fig:datasets}, values of all samples in the dataset are strictly positive.
This is the case to accommodate network models which require positive input values only (common in the field of biologically-plausible networks, as firing rates as well as spike times are typically denoted by positive numbers).
Because of that the yin-yang symbol is not centered around zero.
This however complicates training in neuron models without intrinsic (learnable) bias.
To facilitate training for these models, each sample in the dataset consists not only of the coordinates $(x, y)$ determining the position in the yin-yang symbol but additionally also the values $(1-x, 1-y)$.
This effectively symmetrizes the input and removes the need for a bias even though the yin-yang symbol is not centered around the origin of the coordinate system.

\begin{figure}
    \centering
    \includegraphics[width=0.45\textwidth]{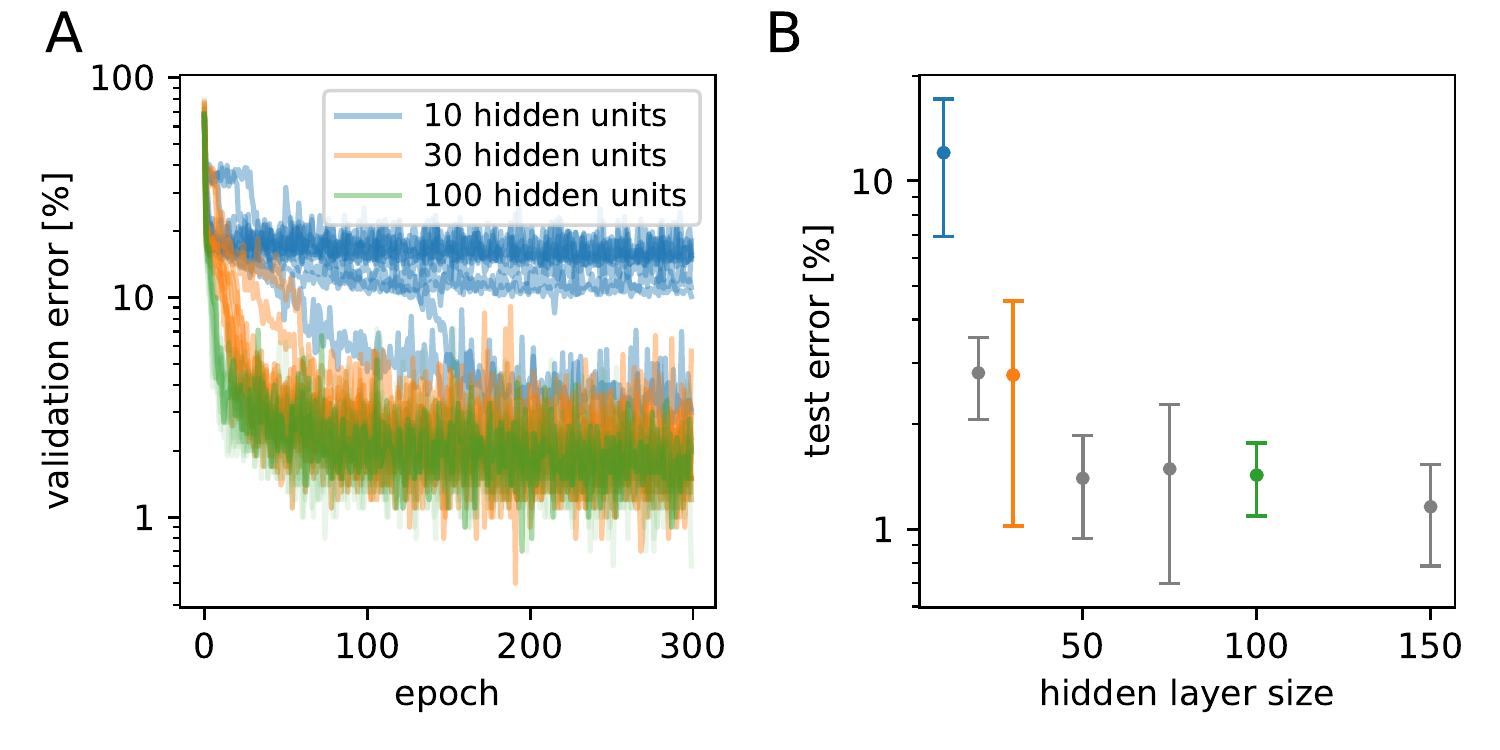}
    \caption{
        \textbf{Impact of hidden layer size on network performance.}
        \textbf{(A)} Validation errors during training for three different network architectures with different hidden layer sizes.
        For each architecture, ten training runs with different random weight initialization are overlaid.
        \textbf{(B)} Mean and standard deviation of the final test error depending on hidden layer size of the network.
        The colored data points correspond to the runs shown in A.
    }
    \label{fig:hidden_sweep}
\end{figure}

\section{Training results}
\begin{table*}[t]
    \caption{
        Mean and standard deviation of the test accuracy for 20 training runs with different random initializations for different network configurations. Training parameters can be found in \cref{tab:train_params}.
    }
    \centering
    \begin{tabular}{l|cc}
        \textbf{network}  & \textbf{hidden layer with 20 neurons} & \textbf{hidden layer with 30 neurons}\\
        \hline
         & & \\[-0.7em]
        deep network & \SI{97.0 \pm 1.6}{\percent} &  \SI{97.6 \pm 1.5}{\percent} \\[0.3em]
        deep network (frozen lower weights)& \SI{78.3 \pm 7.8}{\percent} &  \SI{85.5 \pm 5.8}{\percent} \\[0.3em]
        \hline
         & & \\[-0.9em]
        shallow network & \multicolumn{2}{c}{\SI{63.8 \pm 1.0}{\percent}} \\
    \end{tabular}
    \label{tab:ann_results}
\end{table*}

\begin{table}[!hb]
    \caption{
        Training parameters used to produce the results in \cref{fig:ann_results}.
        \cref{fig:hidden_sweep} uses the same parameters except for the size of the hidden layer.
    }
    \centering
    \begin{threeparttable}
    \begin{tabular}{l|c}
        \textbf{parameter name}  & \textbf{value} \\
        \hline
         & \\[-0.9em]
        activation function & ReLU \\
        size input & $4$ \\
        size hidden layer (for deep net)& $30$ \\
        size output layer & $3$ \\
        training epochs & $300$ \\
        batch size & $20$ \\
        optimizer & Adam, \citep{kingma2014adam}\\
        Adam parameter $\beta$ & $(0.9, 0.999)$ \\
        Adam parameter $\epsilon$ & $10^{-8}$ \\
        learning rate & $0.01$ \\
    \end{tabular}
    \end{threeparttable}
    \label{tab:train_params}
\end{table}

As a baseline for further applications of this dataset we also provide some training results achieved with classical artificial neural networks.
In particular, we compare network performance in three scenarios:
\begin{enumerate}
    \item a network with one hidden layer and fully functional error backpropagation;
    \item a shallow network with only an input and an output layer;
    \item a network with one hidden layer, but with frozen weights between the input and the hidden layer to emulate training with a faulty error backpropagation algorithm.
\end{enumerate}
For all scenarios, we use very small network sizes to emulate a model or hardware prototyping environment.
Incidentally, this is also helpful in highlighting another problem that is frequently overlooked when increasing the network size: because a large enough hidden layer can mask faulty error backpropagation, larger-scale networks are often inadequate for a quantitative verification of credit assignment (precise error propagation) within the studied network model.
This is discussed below in more detail.

The comparison between the three scenarios (\cref{tab:ann_results} and \cref{fig:ann_results}) illustrates a manifest advantage of the Yin-Yang dataset compared to other commonly used datasets of comparable size: both the shallow network and the one with the frozen lower weights are clearly unable to learn the required features to successfully classify the dataset.
This leads to a gap of more than \SI{30}{\percent} between the accuracies achieved by a shallow and a deep network.

The failure of the partially frozen network highlights another important issue for various proposals of bio-plausible solutions to the credit assignment problem.
In large enough networks, the large hidden layers project the input into a very high-dimensional space, which makes classification tasks more easily solvable by the linear classifier embodied by the top layer.
This is commonly referred to as the ``kernel trick'' (see e.g. \cite{scholkopf2001kernel}).
This can easily mask the inability of a network to correctly propagate errors and perform true gradient descent learning.
While this issue would become observable when dealing with more complicated classification problems, it would require using large, deep networks that are not only difficult to debug but, more importantly, would lie beyond the capabilities of typical prototype devices or software simulations.

The Yin-Yang dataset addresses both problems simultaneously, by clearly highlighting faulty error backpropagation already within resource-efficient implementations with hidden layer sizes of around 20 to 30 neurons (see \cref{tab:ann_results}).
Under these circumstances, the difference between the accuracy reached by a properly trained network and the network where only the top weights are trained lies around \SI{20}{\percent} and \SI{12}{\percent} respectively.
This is a much higher gap than in a comparable example with the MNIST dataset, where networks need several hundred hidden neurons to show significant performance improvements beyond linear classifiers \cite{lecun1998gradient}.
However, such sizes automatically introduce the kernel trick: a network with 500 hidden units reaches on average \SI{98.3}{\percent} on MNIST, while the same network with only training in the top layer reaches \SI{94.8}{\percent}.
Unmasking these issues can become crucial in research on biologically plausible forms of credit assignment and (local) synaptic plasticity, where exact error backpropagation is notoriously difficult to realize, both for rate-based models and, even more pronouncedly, for spiking networks.

Another advantage of the Yin-Yang dataset over many other commonly used datasets is the dimensionality of its samples and the network sizes required to learn the task.
Each sample consists of only four input values (compared to, e.g., the 784 input channels required by MNIST), which significantly reduces the required fan-in for hidden neurons.
This can be especially beneficial on neuromorphic platforms, where the number of synaptic connections to a neuron is very often limited by the chip architecture, even more so for early-stage prototypes
(e.g. \cite[Section~3.3]{binas2016precise}, \cite{moradi2013event,schemmel2017accelerated,frenkel20180,nair2019ultra,billaudelle2020versatile}).

Also, this dataset can be learned with a single hidden layer of reasonably small size (\cref{fig:hidden_sweep}).
For consistently high final accuracies, a hidden layer of 20 to 30 neurons is required, but for a small proof-of-concept demonstration of a learning algorithm or hardware prototype, even 10 hidden units are enough to achieve results (around \SI{88}{\percent} accuracy) that would be impossible with shallow networks, or with algorithms that cannot profit from a network's representational hierarchy.
The full set of training parameters can be found in \cref{tab:train_params}.

In addition to the results shown here, the dataset has already been used to showcase algorithms for error backpropagation in spiking neural networks in \citep{goltz2021fast} and \citep{wunderlich2021event}.

\section{Input encoding}
\begin{figure*}[!ht]
    \centering
    \includegraphics[width=0.95\textwidth]{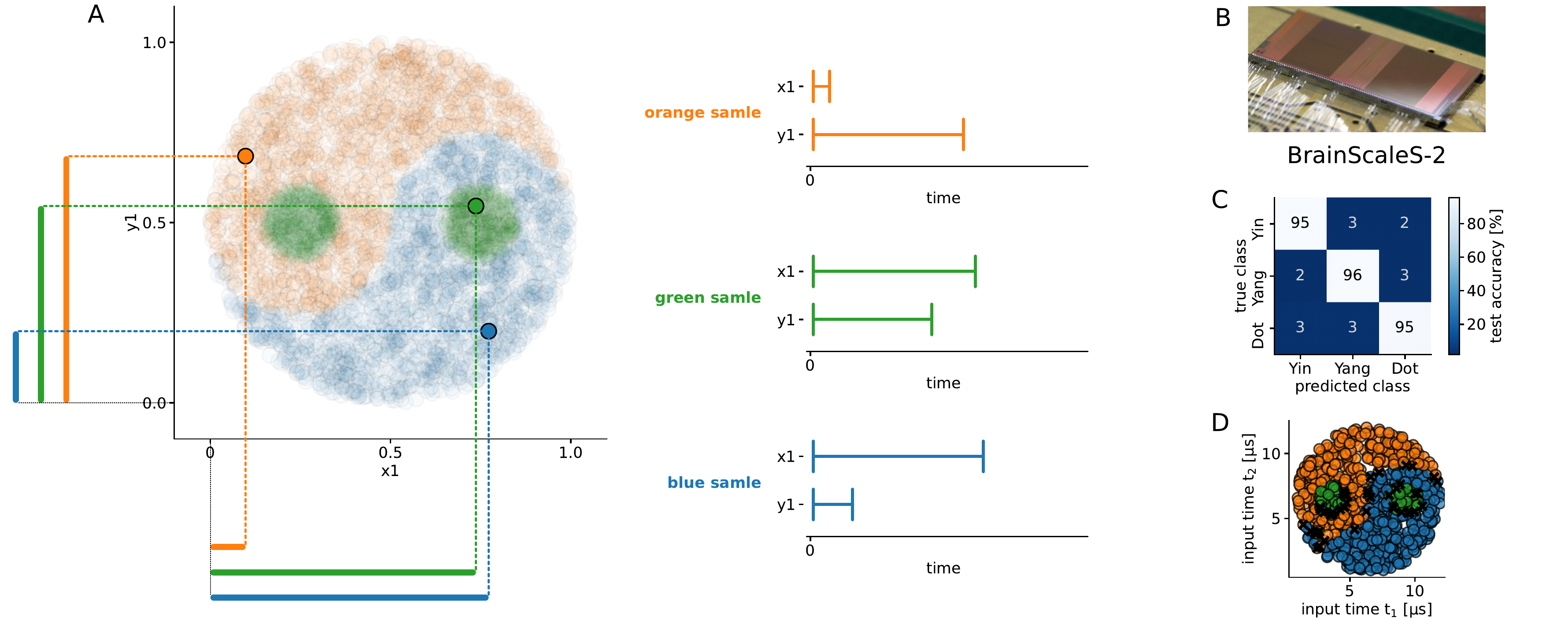}
    \caption{
        \textbf{Spatio-temporal input encoding scheme and classification results on the neuromorphic chip BrainScaleS-2.}
        Panels \textbf{(B-D)} taken from \cite{goltz2019fast}.
        \textbf{(A)} Encoding of the $x,y$-coordinates of the Yin-Yang pattern as input spike times $t_1$ and $t_2$ illustrated on one sample each for the three classes.
        \textbf{(B)} Image of the BrainScaleS-2 ASIC.
        \textbf{(C)} Confusion matrix after training the BrainScaleS-2 chip to classify the Yin-Yang dataset.
        \textbf{(D)} Classification result of the chip on the test set.
        For each input sample the color indicates the class determined by the trained network.
        Wrong classifications are marked with a black X.
        The wrongly classified samples all lie very close to the border between two classes.
    }
    \label{fig:input_encoding}
\end{figure*}

The Yin-Yang dataset can be adapted to suit the needs of very different network models.
Depending on the used network architecture, neuron model and mode of communication between the neurons, different types of information encoding become necessary.
In the following, we discuss several encoding methods that are well-suited for a variety of different network and neuron types.

\subsection{Spatio-temporal input encoding}

Using this dataset for spiking neural networks requires an explicit spatio-temporal input encoding.
In \cite{goltz2021fast} and \cite{wunderlich2021event}, the four input features of the dataset were directly interpreted as the spike times of 4 input neurons (\cref{fig:input_encoding} A).
This was done by choosing parameters $t_\text{early}$ and $t_\text{late}$ as the earliest and latest possible time the input neurons are allowed to spike.
Then the dataset values $\vec{x} = (x, y, 1-x, 1-y)$ were translated into the four spike times $\vec{t} = (t_1, t_2, t_3, t_4)$ as follows:

\begin{align}
    \vec{t} &= t_\text{early} + \vec{x} \cdot \left(t_\text{late} - t_\text{early}\right)
\end{align}
The choice of $t_\text{early/late}$ is dependent on the network architecture and employed learning algorithms.
For \cite{goltz2021fast} it has proven beneficial to choose $t_\text{early}$ slightly after the start of the experiment and $t_\text{late}$ as the sum of the two neuron time constants $t_\text{late} \approx \tau_\text{m} + \tau_\text{syn}$. The classification results achieved with the BrainScaleS-2 chip are shown in \cref{fig:input_encoding}.

Alternatively, a different spike-based spatio-temporal encoding can be achieved implicitly by manipulating input currents, as proposed for example in \cite{cramer2020surrogate}.
Here, each input variable is interpreted as the strength of a constant input current into a leaky-integrate and fire neuron.
The timing of the output spike of the input neurons depends on the strength of the input current $I$ with

\begin{align}
    t_\text{spike} = \tau_{m} \log \frac{I}{I - \theta_\text{I}}
\end{align}
where $\tau_{m}$ denotes the membrane time constant and $\theta_\text{I}$ the minimal current necessary to evoke an output spike.

\subsection{Rate-based input encoding}
Many models for biologically plausible error backpropagation are built around rate-based neuron models (e.g. \cite{sacramento2018,scellier2017equilibrium,haider2021}, for a review see also \cite{whittington2019theories}).
These approaches use continuous rates as an idealized version of rate coding in spiking neurons.
Others build on the same approximations but explicitly use spike-based communication in their neural network implementations (e.g. \cite{schmitt2017neuromorphic,esser2015backpropagation,guerguiev2017towards}).
For such rate-based models, a suitable encoding scheme can be easily realized by designating 4 input neurons and setting their output rates proportional to the values of the respective input feature.

In case of spiking neurons, these four input neurons can produce Poisson spike trains with the same rates as their rate-based counterparts, as, for example, in \cite{schmitt2017neuromorphic}.
Alternatively, regular spike trains could also be used to represent firing rates; while more precise than the intrinsicly stochastic Poisson solution, this scheme has its own potential drawback of making the neuronal input-output function dependent on not just the rate, but also the phase of a neuron's afferents.
Under certain circumstances, encoding an input as a single neuron may not be viable, for example when synaptic bandwidth or neuron firing rate are limited.
In this case, one input can be represented by a population of neurons with a mean firing rate equal to the value of the input.

\FloatBarrier

\subsection*{Code and data availability}\label[methods]{sec:info_codeAv}
Code for the Yin-Yang data set is available at \url{https://github.com/lkriener/yin_yang_data_set}.
The example notebook in the repository includes the plotting of the data samples (\cref{fig:datasets}) and the training of deep and shallow networks (\cref{fig:ann_results}).
Additional data available on request from the authors.

\FloatBarrier
\printbibliography
\addcontentsline{toc}{section}{References}

\section*{Acknowledgment}\label{sec:ack}
We wish to thank Sebastian Billaudelle and Benjamin Cramer for valuable discussions, as well as Mike Davies and Intel for their ongoing support.
We gratefully acknowledge funding from the European Union under grant agreements 604102, 720270, 785907, 945539 (HBP) and the Manfred St{\"a}rk Foundation.

During the development of the dataset some calculations were performed on UBELIX, the HPC cluster at the University of Bern, others were performed on the bwForCluster NEMO, supported by the state of Baden-Württemberg through bwHPC and the German Research Foundation (DFG) through grant no INST 39/963-1 FUGG.
Additionally, our work has greatly benefitted from access to the Fenix Infrastructure resources, which are partially funded from the European Union's Horizon 2020 research and innovation programme through the ICEI project under the grant agreement No. 800858.

\end{document}